\documentclass{article}
\usepackage{spconf,amsmath,graphicx}

\usepackage{balance} 
\usepackage{url}
\usepackage{float}
\usepackage{amsmath,graphicx,amsfonts,amssymb,url}
\usepackage{setspace}
\usepackage[caption=false]{subfig}
\usepackage{blindtext}
\usepackage{enumitem}
\usepackage{color}
\usepackage{multirow}
\usepackage{booktabs}
\newcommand{\squeezeup}{\vspace{-2.5mm}}
\newcommand{\squeezeupf}{\vspace{-1.5mm}}
\definecolor{darkviolet}{rgb}{0.58, 0.0, 0.83}

\newcommand{\suiyi}[1]{\textcolor{black}{#1}}

\newcommand*{\affaddr}[1]{#1} 
\newcommand*{\affmark}[1][*]{\textsuperscript{#1}}

\title{Seeing by haptic glance: reinforcement learning based 3D object Recognition }
\name{Kevin Riou\affmark[1]\thanks{Kevin Riou and Suiyi Ling make equal contribution.}, Suiyi Ling\affmark[1], Guillaume Gallot\affmark[1], Patrick Le Callet\affmark[2]  }\squeezeup 
\address{\affaddr{\affmark[1]CAPACITÉS SAS} \ \ \affaddr{\affmark[2]LS2N, \ University of Nantes} \  \    } 
 \squeezeup  \squeezeup 
\begin{document}
%
\maketitle%
\squeezeup  \squeezeup 
\begin{abstract}
Human is able to conduct 3D recognition by a limited number of haptic contacts between the target object and his/her fingers without seeing the object. This capability is defined as `haptic glance' in cognitive neuroscience. Most of the existing 3D recognition models were developed based on dense 3D data. Nonetheless, in many real-life use cases, where robots are used to collect 3D data by haptic exploration, only a limited number of 3D points could be collected. In this study, we thus focus on solving the intractable problem of how to obtain cognitively representative 3D key-points of a target object with limited interactions between the robot and the object. A novel reinforcement learning based framework is proposed, where the haptic exploration procedure (the agent iteratively predicts the next position for the robot to explore) is optimized simultaneously with the objective 3D recognition with actively collected 3D points. As the model is rewarded only when the 3D object is accurately recognized, it is driven to find the sparse yet efficient haptic-perceptual 3D representation of the object. Experimental results show that our proposed model outperforms the state of the art models. 
  

\end{abstract}
\squeezeup 

\squeezeup \squeezeupf
\section{Introduction}\squeezeup
\label{sec:intro}
\suiyi{With the booming of deep learning, the community has witnessed significant strides in 3D object recognition over the last decade. Most of the existing 3D object recognition models were training on the dataset that consists of dense, clutter-free, canonicalized 3D data. It was proven in a recent study that~\cite{taghanaki2020robustpointset} most of the existing STate-Of-Art (STOA) models, including the ones proposed in~\cite{qi2017pointnet,qi2017pointnet++,uy2019revisiting,taghanaki2020pointmask,liu2019densepoint,li2018pointcnn,wu2019pointconv,liu2019relation,poulenard2019effective,you2018pointwise}, perform significantly poorer on more challenging sets, where the data are with noise, missing parts, sparser points, \textsl{etc}. }
\suiyi{In stark contrast to those models, humans is capable of making reliable decisions with a limited sequence of exploratory movements that provide the most information for the task, regarding what the object is and what to expect from possible exploratory movements based on prior knowledge~\cite{klatzky1995identifying}. This capability of object recognition by a limited number of local tactile cues is defined as `haptic glance'~\cite{rouhafzay2019object,klatzky1999haptic}.  }

\begin{figure}[t]
\subfloat[ ]{\label{plt0_teaster}
\includegraphics[width=0.230\textwidth]{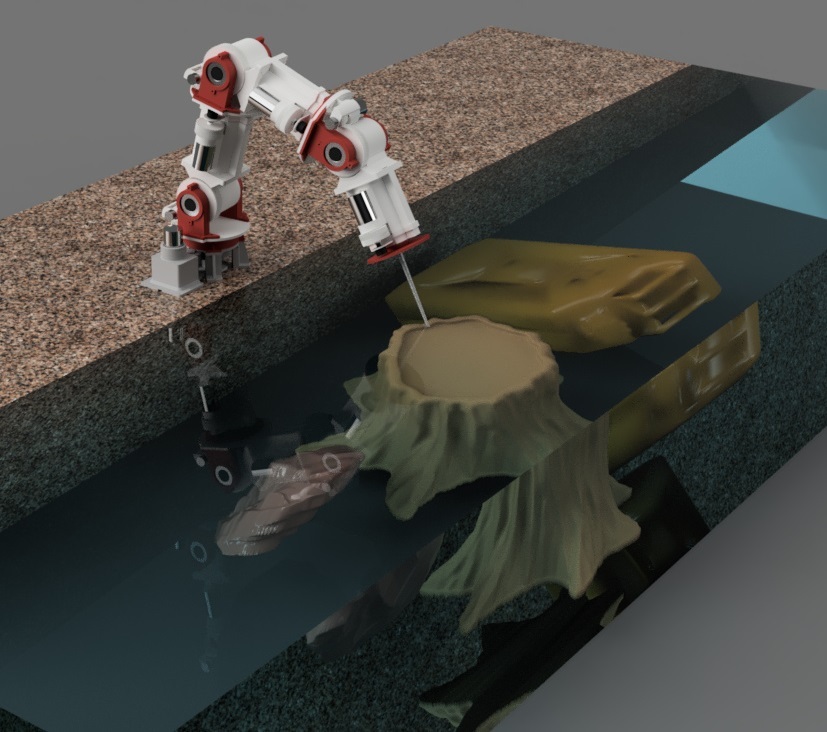}}
\subfloat[ ]{\label{plt1_teaster}\includegraphics[width=0.256\textwidth]{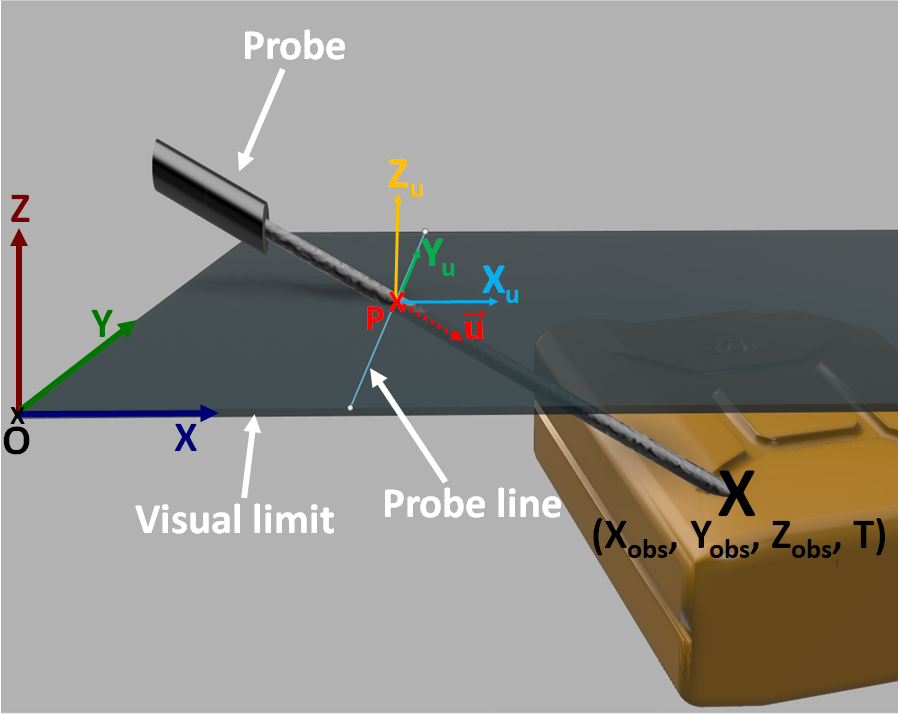}} 
\caption{(a) Use case example; (b) demonstration of reinforcement learning based 3D points collection}
\label{fig:teaser} \squeezeup \squeezeup \squeezeupf
\end{figure}

\suiyi{Robots are envisioned to replace humans for dangerous, inaccessible tasks~\cite{takayama2008beyond}, and are applied in many real-life scenarios. They are of great piratical values, especially for visually unreachable areas exploration, where limited visual information could be obtained. Specifically, robots could be applied to locate, identify, and manipulate/interact with objects under the ground, river/drain (for under-river salvage) \textsl{etc.} with different well-chosen movement schemes as depicted in Fig~\ref{plt0_teaster}. This type of robotic application is known as `haptic exploration'. The fundamental bottleneck within the haptic exploration framework resides in the fact that 1) only a limited number of points could be collected, especially in our case when only one tactile sensor is utilized, which ends out to sparse 3D representations; 2) not all the exploration trials successfully reach a target object, which brings the noise to the collected data; 3) exploration may only provide partial information of the object, resulting in missing objects' parts. In this study, to disentangle the aforementioned obstacles, and further endow the robot with the capability of `haptic glance', we propose a novel reinforcement learning based framework to learn a sparse yet efficient 3D representation.   }

\squeezeup \squeezeup
\section{The Proposed Framework}
\label{sec:pagestyle}
\squeezeup
\subsection{\suiyi{Simulation of Target Use Case Scenario}}
\squeezeup
 \suiyi{Similar to the setup designed in~\cite{Fleer2020}, an in-house simulator was developed to simulate the real-life robot exploration scenario and facilitate the training, testing procedures of the proposed framework. Concretely, a fixed robot was simulated to conduct haptic exploration via sequential tactile probes for 3D recognition to mimic a real-life robotic scenario, where limited visual data could be obtained. An example is depicted in Fig.\ref{plt0_teaster}. Within the simulated environment, 3D objects can be placed and accessed by a robotic hand via haptic probe. The robot hand is equipped with a tactile sensor that measures $U_{len}$, as illustrated in~\ref{plt0_teaster}, to send back pressure-sensation information regarding whether current exploration touches a object or not.} 
  
 \suiyi{Each 3D point is represented by a three-dimensional coordinate $(x, y, z)$, with the origin $O$ of the corresponding coordinate system set as elucidated in Fig.~\ref{plt1_teaster}. Since we aim at recognizing possible invisible objects positioned under the ground, river/drain \textsl{etc.,} each exploration point is constrained to reach the surface of the ground/river, \textsl{i.e.,} $Z = 0$. As such, for each exploration, the starting point is represented by this surface position $P$ and the corresponding orientation $\overrightarrow{u}$. Each $P$ is defined with the coordinate $(P_x, P_y, P_z)$, where $P_{z}=0$, and $\overrightarrow{u}$ with 3 components $(U_x, U_y, U_z)$.} 
 
 \squeezeupf   \squeezeupf
 \begin{figure}[!hbpt]
	\centering
	\includegraphics[width=\columnwidth]{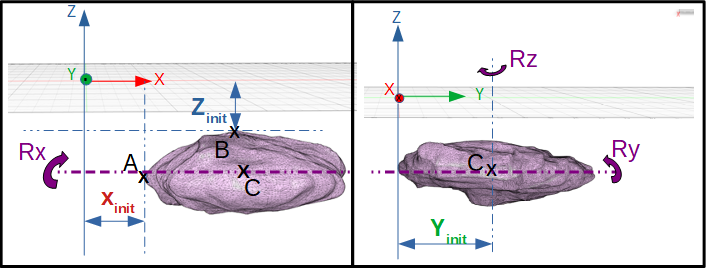}
	\caption{\suiyi{Illustration of objects within the coordinate system.}} \label{fig:objects_init} \squeezeup \squeezeupf  
\end{figure}

 \suiyi{In most of the practical cases, robots scan unexplored areas line-by-line, to ease the transition between the simulation and real-life robotic operation, in this study, haptic exploration was also conducted in a similar fashion. In another word, a certain constant $P_x$ value is first selected, and then the robot hand explores along the $X$ axis progressively. This further defines the line of all possible entry points, \textsl{i.e.,} by projecting $P$ to the $X$ axis, which is named as the `probe line' as highlighted in Fig.~\ref{plt1_teaster}. $U_z$ is constrained to be negative, so that probes only fall in the areas, where possible 3D object is positioned. For each exploratory probe $\{P, \overrightarrow{u}\}$, when the sensor reaches an object, this touched point $(X_{obs}, Y_{obs}, Z_{obs})$ is then calculated within the~\textsl{Ray-casting} system~\cite{pfister1999volumepro} regarding $U_{len}$. In addition, the variable $T$ is set to one, when the sensor touches the object. 3D object could be placed at any initialized position $(X_{init}, Y_{init}, Z_{init})$ with any rotations $R_x, R_y$ and $R_z$, as illustrated in fig.~\ref{fig:objects_init} } \squeezeup

\begin{figure*}[t]
\begin{center}
\includegraphics[width=18cm]{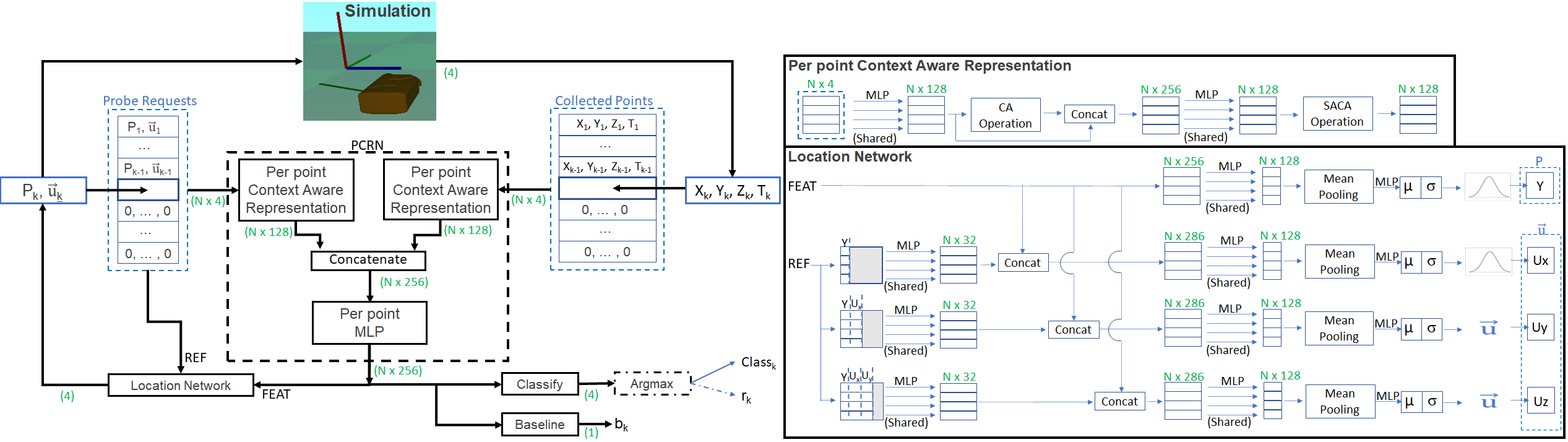}
\end{center}
   \caption{\suiyi{The overall diagram of the proposed reinforcement learning based hepatic exploration and 3D object recognition model.}}
\label{reinforcementFrameworkScheme} \squeezeup  
\end{figure*}

\squeezeup
\subsection{\suiyi{The Reinforcement Haptic Glance Framework}}
\squeezeup

\suiyi{The overall diagram of the proposed reinforcement learning based haptic exploration framework is illustrated in Fig.~\ref{reinforcementFrameworkScheme}. Each probe, \textsl{i.e.}, haptic glance, during the haptic exploration could be parameterized by its position $P$ and orientation $\overrightarrow{u}$. In this study, the haptic control is composed of two parts, including (1) a low-level haptic explorer described by Gaussian distributions $P \backsim \mathcal{N} (\mu_{P}, \sigma_{P}^2)$ and $ \overrightarrow{u}  \backsim \mathcal{N} ( \mu_{\overrightarrow{u}}, \sigma_{\overrightarrow{u}}^2)$, which are parameterized by $\mu_{P}, \sigma_{P},\mu_{\overrightarrow{u}}, \sigma_{\overrightarrow{u}}$; and (2) a high-level reinforcement learner that learns the aforementioned exploration parameters and guarantees the success of 3D recognition. Haptic control and 3D object recognition are achieved via 4 main modules including (a) the Point Cloud Representation Network (PCRN) \suiyi{with} Per Point Context Aware Representation Blocks (P2-CARB), (b) the location network, (c) the classifier, and (d) the reinforcement learning scheme. Details of each module are given below.}

\suiyi{\textbf{(a) The PCRN:} the objective of the framework is to iteratively sample \suiyi{$N$} points \suiyi{via haptic exploration} in order to classify a target object among $M$ possible object categories.} Formally, for each step \suiyi{$k$} between 1 and \suiyi{$N$}, \suiyi{the framework first cast a probe request ($P_k$, $\overrightarrow{u_k}$), and the simulator returns a 3D point ($X_k, Y_k, Z_k, T_k$).} Each new \suiyi{probe request} is stored in \suiyi{a} \suiyi{probe request sequence} \suiyi{$S_R = \{ (P_k, \overrightarrow{u_k}), k \in [1, N] \}$.} \suiyi{Additionally, each} corresponding point \suiyi{sample} is stored in the \suiyi{collected points sequence} \suiyi{$S_C= \{ (X_k, Y_k, Z_k, Tk), k \in [1, N] \}$.} After each \suiyi{probe,} the two sequences are embedded into \suiyi{one} mutual representation space \suiyi{by employing the} P2-CARB. P2-CARB is a permutation invariant network \suiyi{that is} based on \suiyi{the} PointGrow Context Aware (CA) with Self-Attention Context Awareness (SACA) operation \cite{sun2020pointgrow}. \suiyi{In concrete words, after} \suiyi{t}aking \suiyi{$N$} feature vectors as input, CA operation outputs \suiyi{$N$} feature vector\suiyi{. Each output vector contains} information of all the previous collected points aggregated by mean pooling, \suiyi{\textsl{i.e.,}} a permutation invariant operation. SACA is a self-adaptive version of CA, allowing to output, for each point, a weighted aggregation of previous collected point\suiyi{s}. 
The self-attention weights are learned by a Multi-Layer perceptron (MLP), taking a concatenation of point features and context aware features as input. \suiyi{Thus, }SACA-A operation is \suiyi{adapted} in our framework. The P2-CARB is detailed on the right top \suiyi{of Fig.}~\ref{reinforcementFrameworkScheme}.


Each MLP layer represents a set of MLP sharing the same weights\suiyi{, where each MLP processes its own input point independently.}
Each MLP is \suiyi{constructed with} a series of two fully connected layers.
Intermediate\suiyi{ context aware representations of probe requests sequence} and collected points sequence are extracted \suiyi{subsequently} with 2 respective P2-CARBs. These representations are then concatenated and fed into \suiyi{the corresponding} MLP, \suiyi{which} allows to learn \suiyi{the} mutual context aware representation of requests and \suiyi{relevant} \suiyi{pre-}collected points. \suiyi{The whole PCRN process is summarized in Fig. \ref{reinforcementFrameworkScheme}, where each intermediate feature embedding dimensions are \suiyi{highlighted} in green. }\suiyi{The probe request sequence} is of size \suiyi{$N\times4$, and} $P_x$ is constrained to \suiyi{be} a constant \suiyi{while} $P_z$ is constrained to \suiyi{be} 0.~\suiyi{Then, $P$ remains as the only variable to be controlled. Along with the three components of orientation, the request vector is of shape of 4. Taking the mutual representation as input, the classifier seeks to classify the object correctly, and the location network aims at predicting the next location to be probed so that the classification accuracy of the next iteration could be optimized. }




\suiyi{\textbf{(b) The Location Network} is designed based on PointGrow~\cite{sun2020pointgrow} to iteratively predict the next postion to explore by computing the} conditional distribution \suiyi{regarding all the} previous generated points. \suiyi{Particularly}, each new \suiyi{probe request} is generated as a conditional distribution of previous \suiyi{r}equests associated to their corresponding collected points. 
For each \suiyi{probe request}, each component of the request is also generated as a conditional probability of the \suiyi{components of the previous probes} using the masking mechanism illustrated in the right bottom side of \suiyi{Fig.~\ref{reinforcementFrameworkScheme}}: 
\squeezeupf 
\begin{equation} \label{eq:cond_pobs_loc_net}
\begin{split}
\begin{aligned}
  &  P\left( P_{k},\overrightarrow{u_{k}}\right) = P( Py_{k} | \left( S_{R\leq k-1},S_{C\leq k-1}\right))\\
    &\cdot P( Ux_k | \left( S_{R\leq k-1},S_{C\leq k-1}\right), Py_{k})\\
    &\cdot P( Uy_k | \left( S_{R\leq k-1},S_{C\leq k-1}\right), Py_{k}, Ux_k)\\
    &\cdot P( Uz_k | \left( S_{R\leq k-1},S_{C\leq k-1}\right), Py_{k}, Ux_k, Uy_k)\\
\end{aligned}
\end{split} \squeezeup \squeezeup  
\end{equation}

\squeezeup 
\suiyi{As mentioned previously, the low-level haptic explorer is parameterized by $\mu_{P}, \sigma_{P},\mu_{\overrightarrow{u}}, \sigma_{\overrightarrow{u}}$. For the sake of readability, we simply denote $\mu_{P}$ and $\mu_{\overrightarrow{u}}$ as $\mu$, and similarly, $\sigma_{P}$ and $\sigma_{\overrightarrow{u}}$ as $\sigma$.} For each component of the \suiyi{probe request}, \suiyi{$N$} feature vectors are extracted and aggregated by mean pooling, which are used to predict the \suiyi{corresponding} $\mu$ and $\sigma$ parameters. \suiyi{Afterwards, the predicted $\mu$ and $\sigma$ are further activated using the  \textsl{tanh} and \textsl{sigmoid} activation function respectively. Hence, with the $\mu$ and $\sigma$, a stochastic prediction could be made for each component of the next probe request}.

\begin{figure}
	\centering
	\includegraphics[width=0.9\columnwidth]{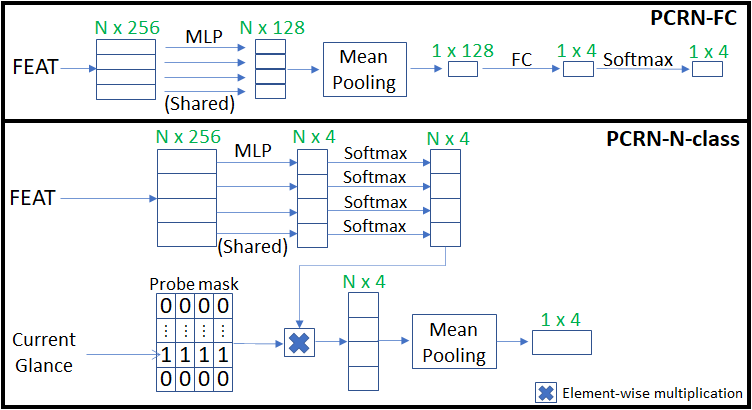} \squeezeupf  
	\caption{The two proposed versions of the classifier} 
	\label{classifier_versions}
	\squeezeup   \squeezeup
\end{figure}

\suiyi{\textbf{(c) The 3D object Classifier:} two} versions of the classifier\suiyi{s} have been developed \suiyi{using the PCRN representation}, as represented in \suiyi{Fig.~}\ref{classifier_versions}. \suiyi{The first module, namely \textsl{PCRN-FC }, aggregates the information of all the probes (all the collected 3D points) by mean pooling and further feed them to a series of two Fully Connected (FC) layers for 3D object classification. }
\suiyi{The \textsl{PCRN-N-class}} version \suiyi{is constituted of} $N$ classifiers, \suiyi{where $N$ is the number of probe/glance performed during exploration. With the designed `probe mask', \textsl{i.e.,} an 1/0 matrix that output only the information of current glance, each classifier is dedicate to classify 3D objects regarding the corresponding probe. }

\suiyi{\textbf{(d) the Reinforcement Learning Scheme:} we adapt the algorithm from~\cite{Willia1992} to train our framework, which} allows \suiyi{the} joint learning of sequential ~\suiyi{haptic exploration} and \suiyi{the efficient sparse 3D} representation \suiyi{of an object by optimizing} the expected cumulative reward $J(\theta)$ :
 \squeezeupf  
\begin{equation} \label{eq:objective_function}
    J(\theta) = E\left[ \sum ^{N-1}_{k=0}r_{k+1} | \pi _{\theta }\right],  \squeezeupf  
\end{equation}
\suiyi{where} $\pi _{\theta }$ is the policy \suiyi{that} predicts the next action to perform $a_k$ \suiyi{regarding} the current state $s_k$. 
\suiyi{In this study, the next action is considered as the next probe request, where the policy is defined based on the Gaussian distributions that specify all the elements of the probe $(P,\overrightarrow{u})$. By this means}, the policy is parameterized directly by $\mu$ and $\sigma$ values outputted by \suiyi{the} location network. Since these values are computed directly from PCRN, the optimization of the policy \suiyi{facilitates the learning of} the mutual representation of probe requests and collected points sequences.
At each time step, the framework tries to classify the object in the ground. It is then rewarded by $r_k = 1 $ if the classification is correct and by $r_k = 0 $ otherwise. By doing so, the point cloud representations is expected to be trained to highlight inter-point relations  that are discriminatory among the available objects classes. \suiyi{With an analogous recipe, the} location network is trained to \suiyi{identify} new points \suiyi{leading to a sparse yet efficient 3D representation} that help in discriminating the object. \suiyi{The} policy gradient $\nabla _{\theta }J(\theta)$ is defined as:  \squeezeup \squeezeup \squeezeup

\begin{equation} \label{eq:policy_gradient} \squeezeupf 
\begin{split}
\begin{aligned}
    \nabla _{\theta }J(\theta) =  \sum ^{N-1}_{k=0} \nabla _{\theta }log(\pi _{\theta }(a_k, s_k))G_k\\
\end{aligned}
\end{split}
\end{equation}

where \suiyi{$ G(k)=\sum ^{N-1}_{k'=k+1}\gamma ^{k'-k+1}r_{k'}$, and} $\gamma$ is the discounted factor \suiyi{that weights} the rewards \suiyi{\textsl{w.r.t}} their time distance from the current state. \suiyi{In this work}, the log-policy gradients for \suiyi{each} \suiyi{probe request component $x$ are given by:} \squeezeup   \squeezeup 

\begin{equation} \label{eq:mu_log_gradient} \squeezeup  
\begin{split}
\begin{aligned}
    \xi _{\mu }=\dfrac{\partial \log  \mathcal{N}\left( x;\mu ,\sigma \right) }{\partial\mu } = \dfrac{x-\mu }{\sigma ^{2}}
\end{aligned}
\end{split} \squeezeup
\end{equation}
\squeezeupf  \squeezeupf  
\begin{equation} \label{eq:sigma_log_gradient}  
\begin{split}
\begin{aligned}
    \xi _{\sigma }=\dfrac{\partial \log  \mathcal{N}\left( x;\mu ,\sigma \right) }{\partial\sigma } = \dfrac{\left( x-\mu \right) ^{2}-\sigma ^{2}}{\sigma ^{3}}
\end{aligned}
\end{split} \squeezeup
\end{equation}

To take the classification task into account in the optimization process, a categorical cross entropy loss is \suiyi{incorporated} to the policy gradient to form a Hybrid Update Rule $\Delta _\theta$ :
\squeezeupf
\begin{equation} \label{eq:update_rule}
\begin{split}
\begin{aligned}
    &\Delta _\theta = -\alpha [\beta  (r_k-b_k)(\xi_{\mu_{Py}} + \xi _{\sigma _{Py}} + \xi _{\mu _{Ux}} + \xi _{\sigma _{Ux}} + \\
    & \xi _{\mu _{Uy}} + \xi _{\sigma _{Uy}} + \xi _{\mu _{Uz}} + \xi _{\sigma _{Uz}} ) + \sum ^{O}_{o=0} y_{o}\log \left( \pi _c \left( o\right) \right)   ]  \\
\end{aligned}
\end{split} \squeezeupf  \squeezeupf 
\end{equation}
 
where $\alpha$ is the learning rate, $\beta$ is a weight balancing \suiyi{the} exploration and classification tasks. \suiyi{$O$ denotes} the set of available objects. $y_o = 1$\suiyi{, if $o$} is the object in the space to explore, $y_o = 0$ otherwise. $ \pi _c (o)$ \suiyi{indicates} the probability that $o$ is the object presented, according to the classifier. $b_k$ is the reinforcement learning baseline\suiyi{~\cite{Willia1992} outputted by the framework as shown in the left-bottom part of Fig.~\ref{reinforcementFrameworkScheme}.} This term reduce\suiyi{s} the \suiyi{variances of} policy gradients variance. 

\squeezeup \squeezeupf
\section{Experiment} \squeezeupf 
\label{sec:Ex} \squeezeupf
\subsection{Experimental setup} \squeezeup

 \begin{figure}[t]
\subfloat[ ]{\label{plt0_objects}
\includegraphics[width=0.22\textwidth]{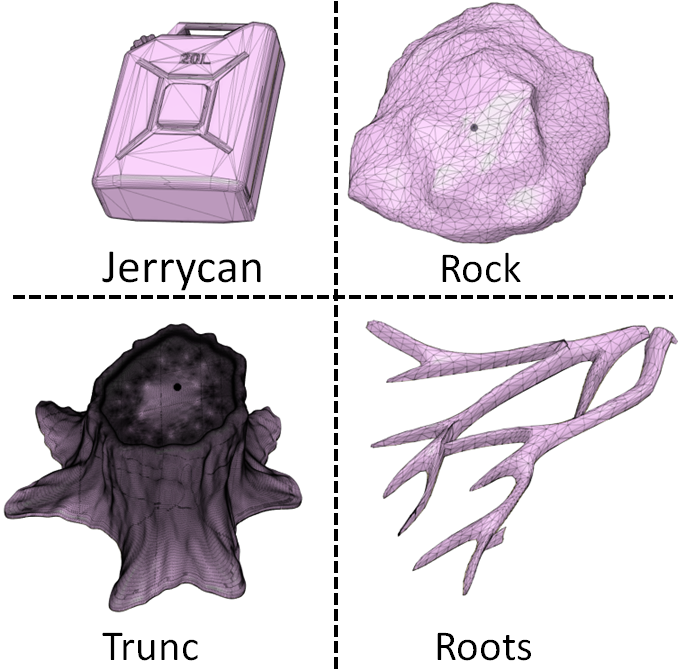}}
\subfloat[ ]{\label{plt1_noise}\includegraphics[width=0.25\textwidth]{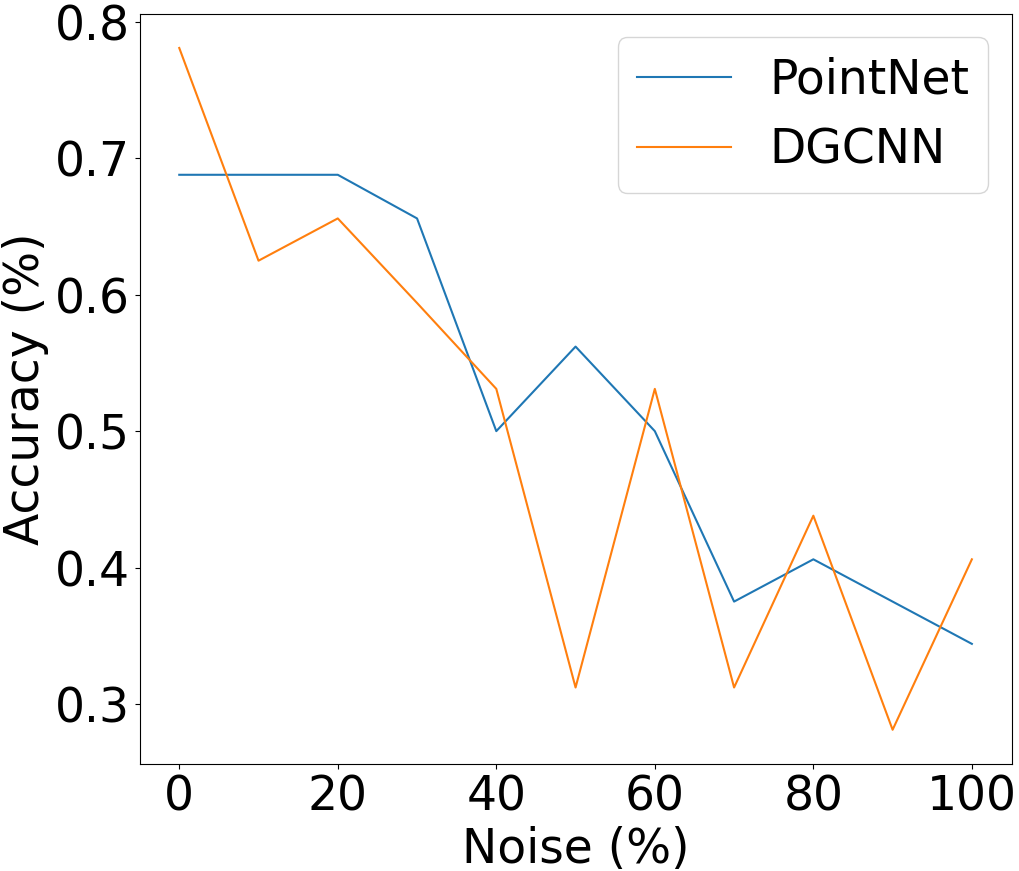}} 
\caption{(a) 4 objects in our dataset; (b) performances of STOA 3D recognition models on sparse, noisy, 3D object sets.  }
\label{fig:teaser} \squeezeup \squeezeup 
\end{figure}

\suiyi{Similar to the setup in~\cite{Fleer2020}, we constructed a 3D dataset with 4 objects that appear frequently in real-life robotic haptic exploration scenario. The four 3D objects are shown in Fig.~\ref{plt0_objects}. }\suiyi{The state-of-the-art haptic shape exploration model~\cite{Fleer2020}  based on LSTM was readjusted to our application (simulator) by adapting the input/output. It is denoted as `LSTM' in this paper, and considered as the baseline model. For fair comparisons, optimized hyper-parameters were employed.} Each \suiyi{model was trained for 8000 steps, where each step is composed of} a batch of 64 objects. 
\suiyi{Every} object was \suiyi{randomly} chosen between the four available categories \suiyi{with equal probability}. The objects were randomly placed in simulation, \suiyi{with random rotations $R_x \in 	\lbrack-10, 10\rbrack$, $R_y \in \lbrack-10, 10\rbrack$  and $R_z \in \lbrack-180, 180\rbrack$,} within an accessible range. Sub-sets of positions/orientations ranges are kept apart \suiyi{as held-out test set} for the evaluation~\cite{Fleer2020}. 


\suiyi{The overall results in terms of averaged accuracy across all the objects are summarized in Table~\ref{table:per_glance_acc}, where the classification with different number of probes were reported. In this work, only 10 probes/glances were considered as done in~\cite{Fleer2020}. As observed, both the proposed \textsl{PCRN-N-class} and PCRN-FC outperform the baseline at the $10_{th}$ probe. PCRN-FC is superior to the other the two models in terms of classification accuracy at each probe. It is showcased that the proposed framework with $N$ classifiers achieves higher accuracy at the final probe by selecting only the top key 3D points with the `probe mask', while the version with sequential FC layers achieves progressively better accuracy at each prob by using all the collected probe, but slightly worse performance at the last glance. They could be employed in different situations. }  \squeezeup   \squeezeupf


\begin{table}[!htpb]
{\small
\begin{center} \squeezeup 
\caption{\suiyi{Performances of haptic exploration models.}}
\begin{tabular}{|c|c|c|c|} \hline
\multicolumn{1}{|c}{\suiyi{Probes}} &
 LSTM~\cite{Fleer2020}  &
 PCRN-N-class &
   PCRN-FC  \\ \hline
        2  & 0.308  & 0.205 & \textbf{0.416} \\
        3  & 0.304  & 0.112 & \textbf{0.505} \\
        4  & 0.289  & 0.222 & \textbf{0.576} \\
        5  & 0.298  & 0.286 & \textbf{0.615} \\
        6  & 0.331  & 0.345 & \textbf{0.649} \\
        7  & 0.526  & 0.429 & \textbf{0.670} \\
        8  & 0.577  & 0.506 & \textbf{0.687} \\
        9  & 0.703  & 0.695 & \textbf{0.710} \\
        10 & 0.723  & \textbf{0.843} & 0.796\\\hline
\end{tabular}\squeezeup  \squeezeup 
\end{center}
\label{table:per_glance_acc}} 
\end{table}
 \squeezeup 
\suiyi{To further evaluate the performances of common 3D object recognition models under a similar haptic exploration scenario, we further construct a noisy sparse 3D objects set (same 4 objects) and tested STOA 3D object recognition models on it. More specifically, each object instance was obtained by randomly sampling 10 points from the original 3D objects with a certain percentage of noise. In this study, `noise' was defined as point that does not fall on the 3D object, to mimic the real-life haptic exploration scenario. As verified in~\cite{taghanaki2020robustpointset}, among the 10 tested STOA 3D recognition models, solely PointNet~\cite{qi2017pointnet} and DGCNN~\cite{wang2019dynamic} were relatively robust under the setting with noise, missing object parts, significantly sparser points \textsl{etc}. Therefore, these two models were tested and the results are reported in Fig.~\ref{plt1_noise}. It is demonstrated that, even though with $0\%$ noise, the performances (acc less than 0.79) of this two models are worse compared to the proposed models. Moreover, their performances drop significantly with the increase of noise rate.}

\squeezeup   \squeezeup 
\section{Conclusion}
\label{sec:Con}  \squeezeup 
In this study, we propose a novel reinforcement learning framework that enables robot for the haptic glance, \textsl{i.e.,} conduct 3D object recognition with sparse yet efficient representation. According to the experimental results, existing 3D object recognition models fail to perform decently for haptic exploration with noisy and sparse 3D data. Conversely, our models achieve decent accuracy and surpass the state-of-the-art haptic exploration model.
 \newpage 
\bibliographystyle{IEEEbib}
\bibliography{refs}

\end{document}